\documentclass[10pt,twocolumn,letterpaper]{article}

\usepackage[pagenumbers]{cvpr}    % To produce the REVIEW version
\usepackage{booktabs}
\usepackage{multirow}
\usepackage{graphicx}
\usepackage[table]{xcolor}
\definecolor{cvprblue}{rgb}{0.21,0.49,0.74}
\usepackage[pagebackref,breaklinks,colorlinks,allcolors=cvprblue]{hyperref}

\def\confName{CVPR}
\def\confYear{2027}

\title{Beyond Trial-and-Error: A Geospatial-Guided Active Reasoning Framework for Robust Aerial Vision-and-Language Navigation}

\author{
Jiawei Ge$^{1,2}$ \quad
Wenkang Gong$^{1}$ \quad
Jiuxin Cao$^{1}$ \quad
Bo Liu$^{1}$ \quad
Lei Cui$^{2}$ \quad
Haoran Wang$^{2}$ \\
Xuelin Zhu$^{3}$ \quad
Biwei Cao$^{4}$\quad
Chang Liu$^{1}$ \quad
Chen Feng$^{5}$ \quad
Ioannis Patras$^{6}$\\[0.6em]
{\small $^{1}$Southeast University}\quad
{\small $^{2}$Institute of AI for Industries, Chinese Academy of Sciences}\\
{\small $^{3}$The Hong Kong Polytechnic University \quad
$^{4}$Hohai University}\\
{\small $^{5}$Queen's University Belfast \quad
$^{6}$Queen Mary University of London}\\[0.4em]
{\tt\small 0jiawei.ge0@gmail.com}
}

\begin{document}
\maketitle
\begin{abstract}
Aerial Vision-and-Language Navigation (VLN) requires UAVs to locate language-specified goals in large-scale urban environments, where objects often appear small and visually ambiguous from high-altitude viewpoints. Existing methods largely follow a reactive paradigm common in indoor VLN, repeatedly updating navigation states from local observations while underutilizing the geospatial structure that remains stable during navigation. As a result, visually similar distractors can cause a reactive goal search to repeatedly reinforce locally plausible goal hypotheses based on passively accumulated evidence, leading to unreliable localization. To address this, we propose GeoReasoner, a robust framework that shifts the paradigm of Aerial VLN from reactive trial-and-error to geospatial-guided active reasoning. Specifically, GeoReasoner first performs pre-search geospatial reasoning to derive region-level priors and constrain the goal hypothesis space. Building on these, it further conducts hypothesis-driven exploration to actively acquire discriminative evidence and refine competing goal hypotheses. Finally, consistency-aware goal verification adversarially assesses complementary evidence sources before stopping, hence reducing premature or erroneous goal confirmations. Experiments on the real-world CityNav benchmark demonstrate that GeoReasoner sets a new SOTA among zero-shot Aerial VLN methods while competing favorably with supervised approaches.
\end{abstract}

\section{Introduction}
\label{sec:intro}

Aerial Vision-and-Language Navigation (Aerial VLN)~\cite{lee2025citynav,liu2023aerialvln} is an emerging embodied navigation task that aims to guide unmanned aerial vehicles (UAVs) to language-specified goals in large-scale outdoor environments. To reach the goal, the UAV must understand geospatial descriptions from an aerial perspective and reason about relations among landmarks, roads, regions, and surrounding contextual objects. This capability is essential for practical UAV applications, including logistics, emergency response, and traffic monitoring~\cite{wang2025openbench,doschl2024say,huang2021decentralized}.

Despite recent progress, existing Aerial VLN methods~\cite{liu2023aerialvln,zhang2025citynavagent,lee2025citynav,cai2025flightgpt,fan2026htnav} \textbf{still largely follow a reactive goal-search paradigm commonly adopted in indoor VLN}~\cite{anderson2018vision,ku2020room}: the UAV agent first observes the current scene, updates its internal navigation state or spatial memory from newly acquired visual evidence, and then selects the next action or waypoint. This paradigm is reasonable in indoor environments, where the navigable space is limited, external spatial priors are often unavailable, and target locations are largely inferred from local observations. However, city-scale aerial scenes are much larger and more open-ended, while high-altitude views make local visual observations more ambiguous. In such settings, the relatively stable geographic structure of urban environments tends to provide more reliable relational cues for goal localization.

Yet reactive navigation driven primarily by local evidence underutilizes these geospatial cues, leaving multiple candidates locally plausible from partial observations and causing goal localization to degenerate into trial-and-error. More importantly, this reactive paradigm relies on evidence passively accumulated along the navigation trajectory, rather than actively exploiting geospatial cues to constrain and compare competing hypotheses. As a result, the agent may become trapped by a locally plausible distractor as seemingly supportive evidence continues to accumulate and reinforce the wrong hypothesis. Such reinforcement can persist even when the hypothesis conflicts with the broader geospatial relations implied by the instruction, eventually leading to premature or erroneous stopping.

\begin{figure}[htbp]
    \centering
    \includegraphics[width=\linewidth]{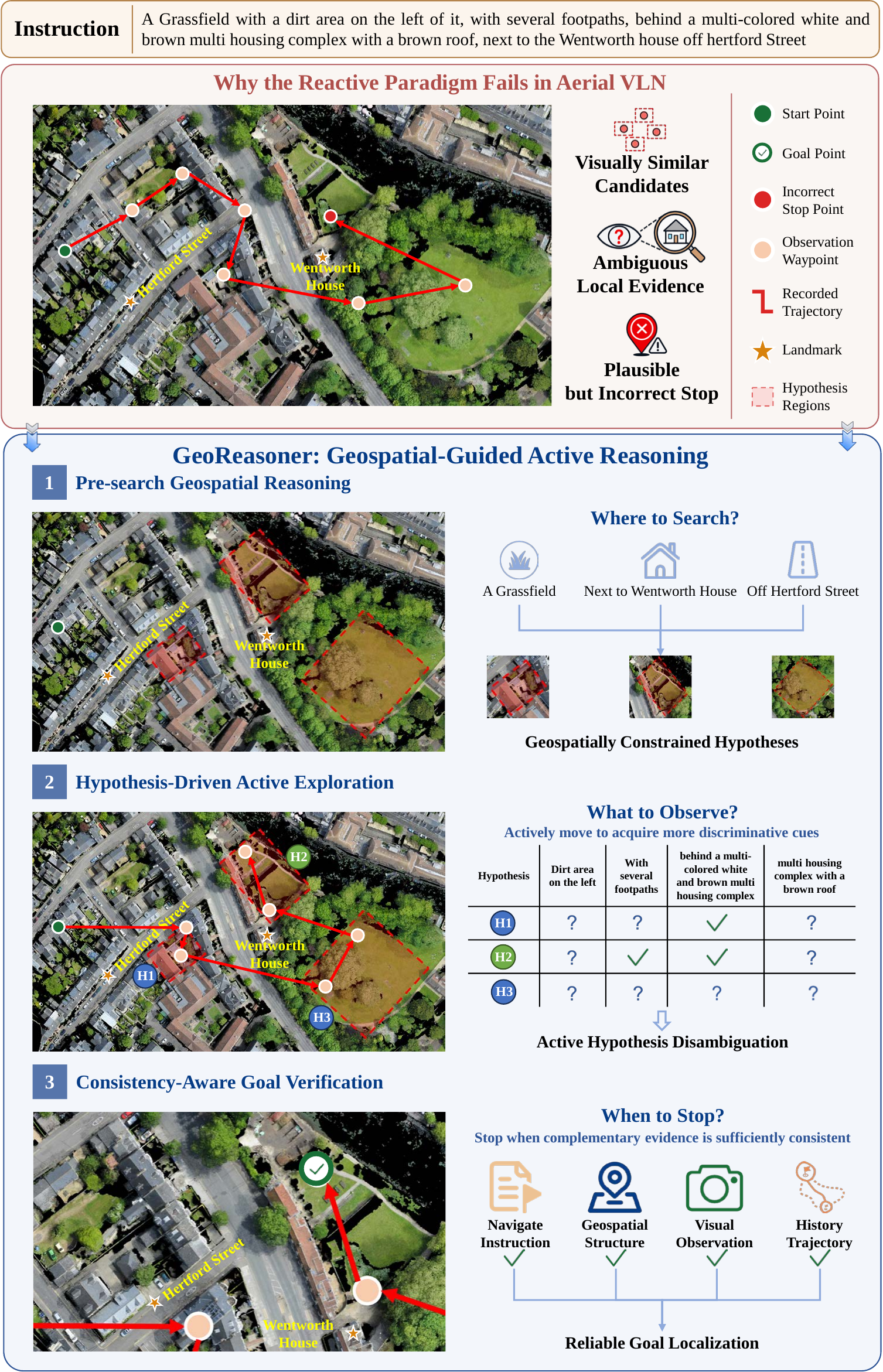}
    \caption{Overview of our framework. Unlike conventional reactive Aerial VLN methods that search for the goal through trial-and-error from passively acquired observations, GeoReasoner exploits geospatial knowledge as a prior before search to constrain the goal hypothesis space, actively acquires discriminative evidence to resolve competing goal hypotheses, and verifies multi-source consistency before stopping.}
    \label{overview}
\end{figure}

To address these issues, we propose a shift from reactive trial-and-error to \textbf{geospatial-guided active reasoning} for Aerial VLN, as shown in Figure \ref{overview}. The key idea is to treat geospatial knowledge \textbf{as an actionable prior} that constrains the goal hypothesis space before searching, guides informative evidence acquisition during navigation, and supports reliable goal verification before stopping. Based on this idea, we introduce \textbf{GeoReasoner}, a geospatial-guided active reasoning framework that organizes goal search in Aerial VLN into three stages: Pre-search Geospatial Reasoning, Hypothesis-Driven Active Exploration, and Consistency-Aware Goal Verification.

Concretely, GeoReasoner first parses language-derived spatial constraints and transforms them into region-level geospatial priors, allowing the UAV to constrain the goal hypothesis space before goal-directed exploration. During navigation, rather than passively evaluating local candidates, the agent performs hypothesis-driven active exploration: it reasons over candidate goal regions, actively seeks discriminative evidence to assess and refine competing hypotheses under geospatial constraints, then progressively localizes the goal through iterative observation and reasoning. Before stopping, GeoReasoner employs an adversarial debate mechanism to verify whether the predicted goal is consistently supported by the instruction, geospatial structure, visual observations, and navigation history. The resulting assessment further informs whether additional exploration or re-planning is necessary. In this way, GeoReasoner establishes a geospatial-guided active reasoning paradigm for Aerial VLN, resolving ambiguous goal hypotheses while improving the reliability of goal localization in complex urban environments.

In summary, our contributions include:
\begin{itemize}

\item We introduce \textbf{GeoReasoner}, a geospatial-guided active reasoning framework that recasts Aerial VLN from reactive trial-and-error into active goal disambiguation, making geospatial knowledge actionable through navigation.

\item We develop the \textbf{Pre-search Geospatial Reasoning} and \textbf{Hypothesis-Driven Active Exploration} schemes, which transform geospatial constraints into competing goal hypotheses and shift goal localization from passive observation accumulation to hypothesis-driven evidence seeking.

\item We introduce a \textbf{Consistency-Aware Goal Verification} mechanism based on adversarial debate that allows stopping when the predicted goal is sufficiently supported by multiple complementary evidence sources, encouraging decisions based on consistent evidence rather than isolated cues.

\item Extensive evaluations on CityNav show that GeoReasoner achieves SOTA \textbf{zero-shot} performance, consistently improving navigation success and localization robustness while narrowing the gap to supervised methods.

\end{itemize}

\section{Related Work}
\label{sec:relate}

\noindent\textbf{Vision-and-Language Navigation (VLN).}
In this task, embodied agents are required to ground natural-language instructions in visual observations and navigate toward specified goals~\cite{anderson2018vision}. Existing methods have progressively improved long-horizon navigation through historical context and structured spatial representations, such as HAMT~\cite{chen2021hamt} and DUET~\cite{chen2022duet}. Recent approaches further leverage large vision-language models to enhance semantic understanding and instruction-guided decision making in navigation~\cite{zhou2024navgpt2,guo2026awarevln,xue2026profocus}. These advances have substantially strengthened instruction grounding, spatial reasoning, and long-horizon decision making in VLN, providing a solid foundation for extending language-guided navigation to aerial platforms.

\noindent\textbf{Aerial Vision-and-Language Navigation.}
Different from conventional VLN, Aerial VLN operates from high-altitude viewpoints over large-scale urban environments, where targets are often visually ambiguous while instructions involve broader geographic relations. AerialVLN~\cite{liu2023aerialvln} first establishes the task in continuous 3D urban environments, while CityNav~\cite{lee2025citynav} further advances toward real-world city-scale navigation, where goals are often described through landmarks and geographic relations. Building on them, subsequent methods progressively strengthen planning and reasoning capabilities. CityNavAgent~\cite{zhang2025citynavagent} and HTNav~\cite{fan2026htnav} strengthen long-horizon planning through hierarchical subgoals and multi-level decision making, respectively. FlightGPT~\cite{cai2025flightgpt} enhances semantic reasoning with vision-language models, and HETT~\cite{ding2026history} incorporates historical context into coarse-to-fine navigation. Despite advances, most Aerial VLN methods still inherit the observation-driven paradigm of indoor VLN, making limited use of relatively stable geospatial information and localizing goals from passively acquired observations. More recent efforts have moved toward exploiting such spatial cues more explicitly: GeoNav~\cite{xu2026geonav} incorporates geographic priors for coarse-to-fine navigation, while LookasideVLN~\cite{ning2026lookasidevln} leverages directional relations for more informed path planning. However, their spatial reasoning mainly determines \emph{where to navigate}, while evidence is obtained as a consequence of navigation rather than deliberately acquired to distinguish competing goal hypotheses. This passive acquisition can expose the UAV agent to accumulated distractors or ambiguous observations, making goal search error-prone. Our proposed GeoReasoner also exploits geospatial priors, but goes beyond navigation assistance by using them to form \emph{competing goal hypotheses}, identify \emph{what evidence is needed} and decide \emph{where to acquire it}.

\noindent\textbf{MLLM-based Spatial Reasoning.}
Recently, MLLMs have progressively evolved from visual understanding toward explicit spatial reasoning. To enhance 3D spatial perception, ShapeLLM~\cite{qi2024shapellm} and SpatialLLM~\cite{ma2025spatialllm} introduce geometry-aware representations and spatially informed designs. Building on this capability, Thinking in Space~\cite{yang2025thinking} and Liu et al.~\cite{liu2025coarse} further strengthen spatial reasoning through explicit cognitive maps and cross-view correspondences. Moving beyond spatial understanding, RingMo-Agent~\cite{hu2026ringmo} extends MLLM reasoning to embodied decision making and long-horizon navigation. Despite these advances, existing benchmarks still reveal limitations in complex spatial relations and orientation~\cite{wang2025spatial457,daxberger2025mmspatial}, highlighting the need for more explicit and reliable spatial reasoning.

\section{Method}
\label{sec:method}

\begin{figure*}[htbp]
    \centering
    \includegraphics[width=\linewidth]{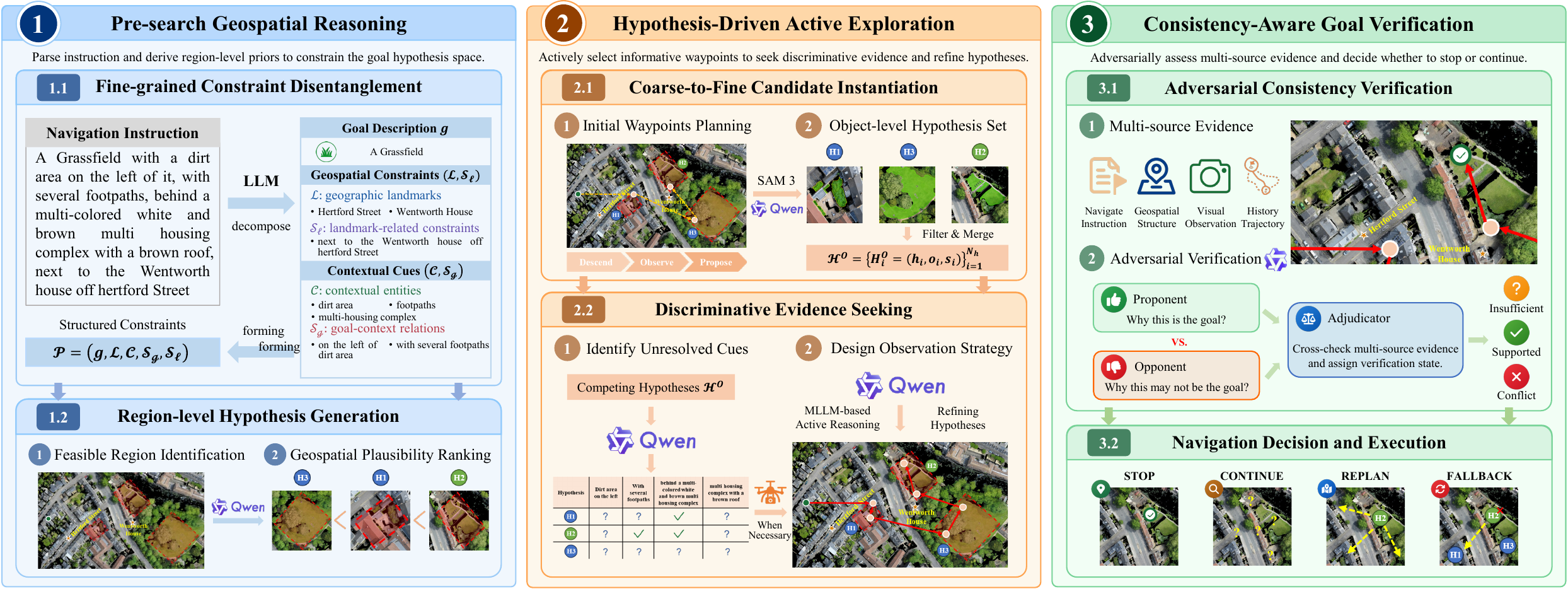}
    \caption{Detailed workflow of our framework. GeoReasoner progressively constrains, refines, and verifies goal hypotheses through three stages: (1) \textbf{Pre-search Geospatial Reasoning} disentangles geographic constraints and contextual cues to generate region-level goal hypotheses. (2) \textbf{Hypothesis-Driven Active Exploration} instantiates object-level hypotheses through initial waypoint exploration and actively seeks discriminative evidence to refine competing hypotheses when necessary. (3) \textbf{Consistency-Aware Goal Verification} employs adversarial reasoning to assess multi-source evidence and guide navigation decisions.}
    \label{framework}
\end{figure*}

\subsection{Overview}
\label{sec:overview}

\noindent\textbf{Problem formulation.}
We follow the standard Aerial VLN setting of CityNav~\cite{lee2025citynav}. Given a natural-language instruction $\mathcal{I}$, a UAV is required to localize the referred goal in a large-scale urban environment.
At each timestep $t$, the agent receives a top-down RGB-D observation
$\mathcal{O}_t=\{\mathcal{O}^{\mathrm{rgb}}_t,\mathcal{O}^{\mathrm{d}}_t\}$
and its current pose
$\mathbf{p}_t=(x_t,y_t,z_t,\theta_t,\psi_t)$,
where $(x_t,y_t,z_t)$ denotes the 3D position and $(\theta_t,\psi_t)$ denote the pitch and yaw angles. The agent selects an action $a_t\in\mathcal{A}$, where
$\mathcal{A}=\{\texttt{forward},\texttt{left},\texttt{right},
\texttt{ascend},\texttt{descend},\texttt{stop}\}$. Additionally, the agent has access to lightweight geospatial information such as geographic entity names and their spatial contours. Since the absolute goal location is unavailable, the agent must infer it from the instruction, geospatial structure, and sequential observations. An episode is considered successful when the agent issues \texttt{stop} within the prescribed distance threshold of the ground-truth goal.

\noindent\textbf{GeoReasoner framework.}
As illustrated in Figure~\ref{framework}, GeoReasoner formulates goal search as a three-stage geospatial-guided active reasoning process.
First, \textit{Pre-search Geospatial Reasoning} (Section~\ref{sec:pre_search}) extracts fine-grained spatial constraints from the instruction and grounds them using the available geospatial information to construct region-level goal hypotheses, before initiating goal-directed exploration. Second, \textit{Hypothesis-Driven Active Exploration} (Section~\ref{sec:active_exploration}) verifies these regions in ascending plausibility order, instantiates representative object-level hypotheses within feasible regions, and actively acquires discriminative evidence for unresolved candidates. Finally, \textit{Consistency-Aware Goal Verification} (Section~\ref{sec:verification})  jointly evaluates linguistic, geospatial, visual, and historical evidence to determine whether to stop, trigger targeted exploration when evidence remains insufficient, or reject conflicting hypotheses and initiate re-planning.

\subsection{Pre-search Geospatial Reasoning}
\label{sec:pre_search}

Navigation instructions often intertwine geographic landmarks with fine-grained contextual cues surrounding the goal. However, these cues differ in spatial granularity, and treating them as a single unit may introduce ambiguity rather than improve goal localization. To address this, GeoReasoner establishes geospatial priors before visual search while preserving contextual cues for subsequent verification. The resulting feasible regions are organized into competing region-level hypotheses to guide active exploration.

\noindent\textbf{Fine-grained Constraint Disentanglement.}
To distinguish geographic constraints and visually grounded contextual relations, we first employ the LLM to decompose the navigation instruction $\mathcal{I}$ into
\begin{equation}
    \mathcal{P}
    =
    \left(
    g,\mathcal{L},\mathcal{C},
    \mathcal{S}_{g},\mathcal{S}_{l}
    \right),
\end{equation}
where $g$ denotes the goal description,
$\mathcal{L}$ represents geographic landmarks, and
$\mathcal{C}=\{c_j\}_{j=1}^{N_c}$ contains individual contextual entities surrounding the goal.
We explicitly preserve the goal--context relations as
\begin{equation}
    \mathcal{S}_{g}
    =
    \left\{
    (g,r_j,c_j)
    \right\}_{j=1}^{N_c},
\end{equation}
where $r_j$ denotes the spatial relation between the goal and contextual entity $c_j$. The remaining landmark-related spatial constraints are represented by
\begin{equation}
    \mathcal{S}_{l}
    =
    \left\{
    (u_k,r_k,v_k)
    \right\}_{k=1}^{K_l}.
\end{equation}
Here, $u_k$ and $v_k$ denote the entities involved in the $k$-th landmark-related constraint, $r_k$ specifies their spatial relation, and $K_l$ is the number of such constraints.

\noindent\textbf{Region-level Hypothesis Generation.}
We then ascend directly from the starting point to altitude \(h_1\) to obtain a high-altitude overview of the surrounding layout, grounding the extracted landmarks in the available geospatial information and translating their spatial relationships into geometric constraints. For instance, \textit{at-corner-of} restricts feasible locations to the vicinity of a specified street intersection, while relations such as \textit{near}, \textit{next-to}, \textit{in-front-of}, and \textit{opposite} further constrain proximity or relative position when their reference entities are geographically grounded. To reduce incorrect pruning caused by ambiguous instructions or imprecise geospatial contours, the LLM further reviews and refines the resulting search regions.

Although geographic constraints narrow the search space, multiple regions may remain plausible due to incomplete or ambiguous spatial relations. Rather than selecting a single region, GeoReasoner maintains competing region-level hypotheses. Specifically, the LLM assesses each retained region against the grounded landmarks and spatial constraints, estimating its geospatial plausibility while identifying supporting evidence and unresolved relations. This yields the region-level hypothesis set
\begin{equation}
    \mathcal{H}^{R}
    =
    \left\{
    H^{R}_j=
    \left(
    \Omega_j,
    s^{R}_j
    \right)
    \right\}_{j=1}^{N_R},
\end{equation}
where $\Omega_j$ denotes the $j$-th feasible region, $s^{R}_j$ represents its geospatial plausibility score, and $N_R$ is the number of retained regions, bounded by \(K_g\). The exploration order jointly considers geospatial plausibility and visiting efficiency. Since coarse geospatial priors may be misleading, GeoReasoner prioritizes less plausible regions to challenge competing hypotheses and avoid premature confirmation.

At the same time, the UAV agent initializes a global memory to retain the supporting evidence and unresolved constraints from $\mathcal{S}_l$, together with goal--context relations from $\mathcal{S}_g$ requiring visual verification. The initial memory is defined as
\begin{equation}
    \mathcal{M}_{init}
    =
    \left\{
    (j,\mathcal{E}^{R}_j,\mathcal{U}^{R}_j)
    \right\}_{j=1}^{N_R},
\end{equation}
where $\mathcal{E}^{R}_j$ contains the geographic evidence supporting $\Omega_j$, and $\mathcal{U}^{R}_j$ records its unresolved spatial constraints and goal--context relations requiring further verification.

\subsection{Hypothesis-Driven Active Exploration}
\label{sec:active_exploration}

While region-level hypotheses identify plausible goal areas, precise localization requires visual grounding and further spatial reasoning. Therefore, within each hypothesis region, GeoReasoner first visits planned waypoints to instantiate object-level goal hypotheses, then employs MLLM reasoning to assess the acquired evidence, identify unresolved cues, and actively guide targeted observations when necessary for hypothesis disambiguation and refinement.

\noindent\textbf{Coarse-to-Fine Candidate Instantiation.}
Within each feasible hypothesis region $\Omega_j$, we descend to altitude \(h_2\) for detailed observation and employ the MLLM (Qwen3.8-Flash \cite{qiu2026design}) to reason over the geographic configuration, instruction-derived spatial constraints, and UAV camera coverage to select a small number of waypoints for initial observation. The UAV then visits these waypoints to acquire visual observations and corresponding poses for subsequent candidate localization.

Afterwards, SAM 3 \cite{carion2026sam} is utilized to extract object masks as proposals from the acquired visual observations, which are further filtered by the MLLM based on the goal appearance described in $g$. The retained proposals are localized using the corresponding observations and UAV poses, and spatially redundant detections are merged into representative candidates using a radius \(r_{\mathrm{merge}}\). Each candidate is then associated with its parent region and relevant instruction-derived spatial constraints, preserving unresolved relations for subsequent active disambiguation. The resulting object-level hypothesis set is defined as
\begin{equation}
    \mathcal{H}^{o}
    =
    \left\{
    H_i^{o}
    =
    (h_i,\mathbf{o}_i,s_i)
    \right\}_{i=1}^{N_h},
\end{equation}
where $h_i=(g,\mathcal{S}_i)$ denotes the semantic-spatial hypothesis, with $\mathcal{S}_i$ representing the goal-context relations requiring verification for candidate $i$. $\mathbf{o}_i$ and $s_i$ denote its estimated geographic location and initial plausibility, respectively. Each hypothesis inherits the geographic constraints of its parent region. Up to $K_h$ representative hypotheses are prioritized for initial detailed inspection, while the remaining candidates are retained.

\noindent\textbf{Discriminative Evidence Seeking.}
Initial observations from a limited number of waypoints may lack sufficient visual and contextual details to distinguish competing goal hypotheses, leaving multiple plausible object-level hypotheses $\mathcal{H}^{o}$ after initial filtering. To address this ambiguity, we apply the MLLM to evaluate these hypotheses against the full instruction, the initial observations, and evidence accumulated in the global agent memory $\mathcal{M}_t$. Candidates with explicit contradictions are eliminated, while the remaining hypotheses are compared to identify unresolved visual and contextual cues that can distinguish them.

When the available evidence is insufficient to resolve these cues, the MLLM determines appropriate observation strategies according to the nature of each cue and its spatial context. Fine-grained target details may require closer or lower-altitude inspection, whereas contextual objects are sought in areas inferred from their spatial relations to the candidate, such as the opposite roadside for a car described as across from the goal. When additional observations are needed, the MLLM combines these observation requirements with the current UAV pose and exploration history to actively select targeted observation waypoints that can help distinguish competing hypotheses. When multiple supplementary waypoints are identified, they are organized into an efficient exploration route by solving a Traveling Salesman Problem (TSP) \cite{applegate2011traveling}. The UAV then executes navigation actions $a_t\in\mathcal{A}$ toward the selected waypoints, acquiring additional observations $O_{t+1}$ and corresponding poses $p_{t+1}$ of the candidate objects or their surrounding areas. Supplementary observation attempts are limited to a total budget \(B_v\) per episode.

The MLLM then organizes evidence from the initial and, when available, supplementary observations into a structured record for each relevant hypothesis:
\begin{equation}
    \mathcal{B}_i^t
    =
    (n_i^t,u_i^t,c_i^t),
\end{equation}
where $n_i^t$ summarizes the evidence assessed at step $t$, $u_i^t$ records the cues that remain unresolved, and $c_i^t$ indicates whether the evidence supports, contradicts, or suggests revising the hypothesis. The resulting record is incorporated into the global agent memory:
\begin{equation}
    \mathcal{M}_{t+1}
    =
    \operatorname{Update}
    \left(
    \mathcal{M}_t,\mathcal{B}_i^t
    \right).
\end{equation}
Based on the updated memory, GeoReasoner refines candidate locations using available observations and their corresponding UAV poses when more accurate estimates become available. The accumulated visual and contextual evidence further guides hypothesis re-ranking and the elimination of inconsistent candidates. The refined hypothesis set is represented as
\begin{equation}
    \hat{\mathcal{H}}^{o}
    =
    \left\{
    \hat{H}_i^{o}
    =
    (h_i,\hat{\mathbf{o}}_i,\hat{s}_i)
    \right\}_{i=1}^{N_h'},
\end{equation}
where $\hat{\mathbf{o}}_i$ and $\hat{s}_i$ denote the refined location and updated plausibility, respectively, and $N_h'$ is the number of surviving hypotheses. The refined hypotheses $\hat{\mathcal{H}}^{o}$ and accumulated agent memory $\mathcal{M}_T$ are subsequently passed to consistency-aware goal verification for the final navigation decision.

\subsection{Consistency-Aware Goal Verification}
\label{sec:verification}
Even after active evidence seeking and hypothesis refinement, the highest-ranked candidate may lack decisive evidence distinguishing it from competing alternatives. Hence, we perform a final check to assess its consistency with the instruction-specified visual attributes and spatial relations, and determine whether the accumulated evidence is sufficient for goal confirmation.

\noindent\textbf{Adversarial Consistency Verification.}
Given the refined hypotheses $\hat{\mathcal{H}}^{o}$ and global agent memory $\mathcal{M}_T$, we introduce a single-round debate among three role-specific MLLM agents to verify the leading hypothesis against competing alternatives. Specifically, the \emph{proponent} constructs a supporting argument by identifying visual, contextual, and geospatial evidence consistent with the navigation instruction. Conversely, the \emph{opponent} challenges this argument by identifying potential contradictions and examining alternative hypotheses that could also explain the observed evidence. 

The \emph{adjudicator} then evaluates both arguments by examining whether the visual attributes and contextual relations of each hypothesis are jointly consistent with the instruction, geographic constraints, and observations accumulated throughout exploration, while leveraging the navigation history to identify supporting or contradictory evidence from previously visited locations. Based on this consistency assessment, it rules out competing hypotheses that contradict verified evidence while retaining those whose distinguishing cues remain unresolved. This process prevents the leading hypothesis from being accepted solely on the basis of its relative plausibility.

For each hypothesis $\hat{H}_i^{o}\in\hat{\mathcal{H}}^{o}$, the adjudicator summarizes supporting, contradictory, and unresolved evidence in a structured assessment $\mathcal{E}_i$, and assigns a verification state:
\begin{equation}
    y_i \in
    \{
    \texttt{supported},
    \texttt{insufficient},
    \texttt{conflict}
    \},
\end{equation}
where \texttt{supported} indicates sufficient evidence consistency with no substantial competing ambiguity, \texttt{insufficient} denotes unresolved evidence or viable alternatives, and \texttt{conflict} indicates an explicit violation of essential constraints.

\noindent\textbf{Navigation Decision and Execution.}
The adjudicator's verdict directly guides the final navigation decision. A supported hypothesis is accepted as the goal, while targeted evidence seeking is resumed only when unresolved cues could alter the decision. If further informative observations are infeasible, GeoReasoner performs a budget-constrained fallback selection among the remaining non-conflicting hypotheses, rather than treating insufficient evidence as verified goal confirmation. When all candidates are rejected, re-planning is triggered to revisit region-level hypotheses using accumulated evidence, allowing previously excluded regions to be reconsidered.

Let $\mathcal{H}_{\mathrm{surv}}=\{\hat{H}_i^{o}\mid y_i\neq\texttt{conflict}\}$ denote the surviving hypotheses. When fallback selection is required, the agent performs evidence voting over the adjudicator's existing assessments:
\begin{equation}
    H^{\mathrm{vote}}
    =
    \operatorname{Vote}
    \left(
    \left\{
    (\hat{H}_i^{o},\mathcal{E}_i)
    \right\}_{\hat{H}_i^{o}\in\mathcal{H}_{\mathrm{surv}}}
    \right).
\end{equation}

Let $\eta_t\in\{0,1\}$ indicate whether further informative exploration has been exhausted. The navigation decision follows the ordered rules:
\begin{equation}
    d_t =
    \begin{cases}
        \texttt{STOP}(H^\star), &
        \exists\,i:\ y_i=\texttt{supported},
        \\[2mm]
        \texttt{REPLAN}, &
        \mathcal{H}_{\mathrm{surv}}=\emptyset,
        \\[2mm]
        \texttt{EXPLORE}, &
        \eta_t=0,
        \\[2mm]
        \texttt{STOP}(H^{\mathrm{vote}}), &
        \text{otherwise},
    \end{cases}
\end{equation}
where $H^\star$ denotes the hypothesis confirmed by the adjudicator, and $\mathrm{STOP}(H)$ denotes navigating to the estimated location of $H$ before issuing the stop action. The number of additional re-planning attempts is limited to \(R_{\max}\). If all hypotheses remain conflicting after exhausting the re-planning budget, the episode terminates without a verified goal. By evaluating competing hypotheses through adversarial reasoning and consistency checks, GeoReasoner mitigates premature goal commitment while preserving the efficiency of targeted exploration.

\section{Experiments}
\label{sec:exps}

\subsection{Experimental Setup}

\noindent\textbf{Dataset.}
We conduct experiments on CityNav~\cite{lee2025citynav}, a real-world aerial VLN benchmark containing 32,637 instruction-trajectory pairs for 5,850 target objects across Cambridge and Birmingham, covering 4.65~km$^2$. The dataset comprises 22,002 training, 2,498 validation-seen, 2,826 validation-unseen, and 5,311 test-unseen episodes. Following~\cite{fan2026htnav}, we further evaluate navigation performance across three difficulty levels (Easy, Medium, and Hard) based on the start-to-goal distance.

\noindent\textbf{Evaluation Metrics.}
Following CityNav~\cite{lee2025citynav}, we adopt four metrics: Navigation Error (NE), Success Rate (SR), Oracle Success Rate (OSR), and Success weighted by Path Length (SPL). NE measures the final distance to the goal, while SR and OSR evaluate whether the UAV reaches within 20\,m of the goal at termination or during navigation, respectively. SPL further accounts for navigation efficiency by incorporating path length into task success.

\begin{table*}[ht]
\centering
\caption{Comparison with state-of-the-art methods on the original CityNav benchmark. NE is measured in meters, while SR, OSR, and SPL are reported as percentages. Best and second-best results are highlighted in \textbf{bold} and \underline{underline}, respectively.}
\label{tab:citynav_sota}
\setlength{\tabcolsep}{2.9pt}
\renewcommand{\arraystretch}{1.10}
\resizebox{\textwidth}{!}{%
\begin{tabular}{lc*{12}{c}}
\toprule
\multirow{2}{*}{Method} &
\multirow{2}{*}{Learning Setting} &
\multicolumn{4}{c}{Test-Unseen} &
\multicolumn{4}{c}{Val-Unseen} &
\multicolumn{4}{c}{Val-Seen} \\
\cmidrule(lr){3-6}
\cmidrule(lr){7-10}
\cmidrule(lr){11-14}
& & NE$\downarrow$ & SR$\uparrow$ & OSR$\uparrow$ & SPL$\uparrow$
  & NE$\downarrow$ & SR$\uparrow$ & OSR$\uparrow$ & SPL$\uparrow$
  & NE$\downarrow$ & SR$\uparrow$ & OSR$\uparrow$ & SPL$\uparrow$ \\
\midrule

Seq2Seq+GSM & \multirow{10}{*}{Supervised}
& 98.10 & 3.81 & 13.92 & 2.79
& 78.60 & 5.13 & 10.90 & 4.65
& 58.50 & 8.43 & 17.31 & 7.28 \\

CMA+GSM &
& 94.6 & 4.68 & 12.01 & 4.05
& 75.9 & 4.38 & 9.29 & 3.90
& 68.0 & 6.25 & 13.28 & 5.40 \\

AerialVLN+GSM &
& 85.1 & 6.72 & 18.21 & 5.16
& 72.7 & 6.35 & 15.24 & 5.06
& 56.6 & 10.16 & 22.20 & 7.89 \\

MGP &
& 93.80 & 6.38 & 26.04 & 6.08
& 75.10 & 5.84 & 22.19 & 5.56
& 59.70 & 8.69 & 35.51 & 8.28 \\

FlightGPT &
& 76.20 & 21.20 & 35.38 & 19.24
& 68.10 & 14.69 & 29.33 & 13.24
& 66.10 & 17.57 & 30.26 & 15.78 \\

FlightGPT+GPOD &
& \underline{54.85} & \underline{29.15}
& \underline{44.94} & \underline{26.85}
& 61.88 & \underline{24.77}
& \underline{39.14} & \underline{22.75}
& 61.56 & 23.86 & 36.71 & 21.83 \\

RingMo-Agent &
& 64.2 & 21.15 & 33.79 & 20.92
& 61.6 & 16.99 & 30.11 & 16.66
& \textbf{35.7} & \textbf{45.81}
& \underline{57.07} & \textbf{45.59} \\

HETT &
& 72.90 & 22.97 & 39.30 & 17.01
& 62.10 & 17.48 & 25.09 & 14.46
& 45.20 & 25.16 & 48.40 & 23.01 \\

HTNav &
& 68.50 & 22.23 & 41.02 & 19.63
& \underline{59.80} & 15.85 & 34.43 & 13.55
& 47.20 & 28.30 & 47.92 & 25.07 \\

DroneNav &
& 74.9 & 22.47 & 30.22 & 20.38
& 63.5 & 15.06 & 25.81 & 13.27
& 54.3 & 18.94 & 36.65 & 17.37 \\

\midrule

GeoNav &
& 73.5 & 25.9 & 41.6 & 16.0
& 64.1 & 16.9 & 35.5 & 9.9
& 58.6 & 21.8 & 39.9 & 12.3 \\

\rowcolor{gray!20}
\textbf{GeoReasoner} & \multirow{-2}{*}{Zero-shot}
& \textbf{38.97} & \textbf{37.77}
& \textbf{59.63} & \textbf{27.85}
& \textbf{49.60} & \textbf{35.14}
& \textbf{61.36} & \textbf{26.57}
& \underline{43.07} & \underline{35.27}
& \textbf{62.77} & \underline{26.93} \\

\bottomrule
\end{tabular}%
}
\vspace{-2mm}
\end{table*}

\begin{table*}[ht]
\centering
\caption{Navigation performance across different difficulty levels on the original CityNav Test-Unseen split. Best and second-best results are highlighted in \textbf{bold} and \underline{underline}, respectively.}
\label{tab:citynav_difficulty}
\setlength{\tabcolsep}{3.4pt}
\renewcommand{\arraystretch}{1.10}
\resizebox{\textwidth}{!}{%
\begin{tabular}{lc*{12}{c}}
\toprule
\multirow{2}{*}{Method} &
\multirow{2}{*}{Learning Setting} &
\multicolumn{4}{c}{Easy} &
\multicolumn{4}{c}{Medium} &
\multicolumn{4}{c}{Hard} \\
\cmidrule(lr){3-6}
\cmidrule(lr){7-10}
\cmidrule(lr){11-14}
& & NE$\downarrow$ & SR$\uparrow$ & OSR$\uparrow$ & SPL$\uparrow$
  & NE$\downarrow$ & SR$\uparrow$ & OSR$\uparrow$ & SPL$\uparrow$
  & NE$\downarrow$ & SR$\uparrow$ & OSR$\uparrow$ & SPL$\uparrow$ \\

\midrule

Seq2Seq & \multirow{4}{*}{Supervised}
& 238.80 & 3.07 & 14.70 & 2.64
& 246.50 & 0.43 & 3.87 & 0.38
& 253.10 & 0.48 & 4.38 & 0.44 \\

CMA &
& 260.70 & 0.49 & 16.69 & 0.44
& 241.20 & 1.10 & 7.67 & 1.09
& 253.80 & 0.96 & 1.64 & 0.95 \\

MGP &
& 98.90 & 6.15 & 39.89 & 5.48
& 90.90 & 6.29 & 21.47 & 6.21
& 90.00 & 6.80 & 12.10 & 6.78 \\

HTNav &
& 68.90 & 19.99 & \underline{45.18} & 14.94
& 68.20 & 20.40 & 39.44 & \underline{18.69}
& \underline{68.10} & \underline{21.91}
& \underline{34.14} & \underline{21.11} \\

\midrule

GeoNav &
& \underline{66.1} & \underline{26.1}
& 40.1 & \underline{15.2}
& \underline{53.8} & \underline{22.9}
& \underline{39.6} & 17.1
& 68.9 & 16.7 & 22.9 & 12.5 \\

\rowcolor{gray!20}
\textbf{GeoReasoner} & \multirow{-2}{*}{Zero-shot}
& \textbf{36.63} & \textbf{39.40}
& \textbf{64.27} & \textbf{24.75}
& \textbf{39.65} & \textbf{37.84}
& \textbf{60.68} & \textbf{29.21}
& \textbf{41.43} & \textbf{35.45}
& \textbf{52.13} & \textbf{30.59} \\

\bottomrule
\end{tabular}%
}
\vspace{-2mm}
\end{table*}

\begin{table*}[ht]
\centering
\caption{Navigation results on the Hard subsets of CityNav. Best and second-best results are highlighted in \textbf{bold} and \underline{underline}, respectively.}
\label{tab:citynav_hard_repro}
\setlength{\tabcolsep}{3.1pt}
\renewcommand{\arraystretch}{1.10}
\resizebox{\textwidth}{!}{%
\begin{tabular}{lc*{12}{c}}
\toprule
\multirow{2}{*}{Method} &
\multirow{2}{*}{Learning Setting} &
\multicolumn{4}{c}{Test-Unseen-Hard} &
\multicolumn{4}{c}{Val-Unseen-Hard} &
\multicolumn{4}{c}{Val-Seen-Hard} \\
\cmidrule(lr){3-6}
\cmidrule(lr){7-10}
\cmidrule(lr){11-14}
& & NE$\downarrow$ & SR$\uparrow$ & OSR$\uparrow$ & SPL$\uparrow$
  & NE$\downarrow$ & SR$\uparrow$ & OSR$\uparrow$ & SPL$\uparrow$
  & NE$\downarrow$ & SR$\uparrow$ & OSR$\uparrow$ & SPL$\uparrow$ \\
\midrule

HETT & \multirow{2}{*}{Supervised}
& \underline{45.14} & \underline{24.00}
& \underline{42.51} & 21.97
& \underline{58.84} & \underline{18.87}
& \underline{32.01} & \underline{17.04}
& \textbf{41.40} & \underline{27.95}
& \underline{43.74} & \underline{25.89} \\

FlightGPT &
& 73.01 & 23.09 & 35.51 & \underline{22.05}
& 69.92 & 14.99 & 23.22 & 14.20
& 71.97 & 14.62 & 25.50 & 13.90 \\

\midrule

GeoNav (Qwen3.8-Flash) &
& 59.90 & 12.16 & 25.38 & 10.59
& 77.87 & 9.71 & 21.09 & 8.54
& 71.87 & 8.54 & 19.77 & 7.00 \\

\rowcolor{gray!20}
\textbf{Ours} (Qwen3.8-Flash) & \multirow{-2}{*}{Zero-shot}
& \textbf{41.43} & \textbf{35.45}
& \textbf{52.13} & \textbf{30.59}
& \textbf{54.32} & \textbf{34.51}
& \textbf{55.78} & \textbf{29.79}
& \underline{47.79} & \textbf{33.10}
& \textbf{54.74} & \textbf{28.21} \\

\bottomrule
\end{tabular}%
}
\vspace{-2mm}
\end{table*}

\noindent\textbf{Comparison Methods.}
We compare GeoReasoner with 11 aerial VLN baselines, including conventional observation-driven methods (Seq2Seq+GSM~\cite{krantz2020beyond,lee2025citynav}, CMA+GSM~\cite{krantz2020beyond,lee2025citynav}, AerialVLN+GSM~\cite{liu2023aerialvln,lee2025citynav}, and MGP~\cite{lee2025citynav}), recent learning-based approaches (FlightGPT~\cite{cai2025flightgpt}, RingMo-Agent~\cite{hu2026ringmo}, HETT~\cite{ding2026history}, HTNav~\cite{fan2026htnav}, and DroneNav~\cite{liu2026dronenav}), and geographically informed methods (FlightGPT+GPOD~\cite{cai2025flightgpt,liu2026gpod} and GeoNav~\cite{xu2026geonav}).

\noindent\textbf{Implementation Details.}
GeoReasoner operates in a zero-shot setting, employing Qwen3.8-Flash \cite{qiu2026design} for multimodal reasoning and SAM~3~\cite{carion2026sam} for object proposal generation. For pre-search stage, we retain at most $K_g=6$ feasible regions and perform an initial high-altitude observation at $h_1=100$\,m. For active exploration stage, detailed observations are conducted at $h_2=50$\,m, with spatially redundant SAM~3 proposals merged within $r_{\mathrm{merge}}=9$\,m. Up to $K_h=9$ representative candidates are selected for initial inspection, and supplementary observations are limited to $B_v=4$ attempts per episode. For goal verification stage, we employ a single-round adversarial debate, allowing at most $R_{\max}=2$ additional re-planning attempts.

\subsection{Experimental Results}

\noindent\textbf{Quantitative Comparisons.}
Table~\ref{tab:citynav_sota} compares GeoReasoner with supervised and zero-shot methods on CityNav. \textbf{Despite requiring no task-specific training, GeoReasoner outperforms all methods across all metrics on unseen splits}, including supervised ones. On Val-Seen, GeoReasoner achieves the highest OSR of 62.77\%, while RingMo-Agent leads in the remaining metrics, potentially benefiting from task-specific training on seen environments.

Compared with reactive navigation methods, ranging from conventional ones (Seq2Seq+GSM, CMA+GSM, and MGP) to recent reasoning-based methods (FlightGPT, FlightGPT+GPOD, HETT, and HTNav), GeoReasoner achieves substantial improvements. Across both unseen splits, GeoReasoner surpasses the best supervised methods, reducing NE by up to 15.88\,m while improving SR, OSR, and SPL by up to 10.37, 22.22, and 3.82 points, respectively. These consistent gains, achieved without task-specific training, highlight the superiority of geospatial-guided active reasoning over reactive navigation paradigms.

Compared with the zero-shot GeoNav, which leverages geospatial priors to guide navigation but still acquires evidence as a consequence of exploration, GeoReasoner reduces Test-Unseen NE by 34.53\,m while improving SR, OSR, and SPL by 11.87, 18.03, and 11.85 points, respectively. This substantial improvement highlights the advantage of shifting from passive exploration to hypothesis-driven active reasoning, where discriminative observations are deliberately acquired to resolve competing hypotheses.

\noindent\textbf{Robustness across Navigation Difficulties.}
We also evaluate GeoReasoner's robustness under increasing navigation difficulty. As shown in Table~\ref{tab:citynav_difficulty}, GeoReasoner achieves the best results across all metrics on Easy, Medium, and Hard episodes. Notably, its SR decreases by only 3.95 points from Easy to Hard, compared with a 9.40-point drop for GeoNav. On Hard episodes, GeoReasoner achieves 35.45\% SR and 52.13\% OSR, outperforming HTNav by 13.54 and 17.99 points, respectively.

We further examine robustness across the three Hard evaluation splits in Table~\ref{tab:citynav_hard_repro}. Our zero-shot method, GeoReasoner, consistently surpasses the supervised HETT and FlightGPT in SR, OSR, and SPL. Moreover, using the same MLLM (Qwen3.8-Flash) as the reproduced GeoNav, GeoReasoner still achieves SR improvements of 23.29, 24.80, and 24.56 points across the three splits, respectively. These results confirm the robustness of our geospatial-guided active reasoning framework under challenging conditions.

\begin{table}[ht]
\centering
\caption{Ablation results of GeoReasoner on Test-Unseen.}
\label{tab:citynav_ablation}
\setlength{\tabcolsep}{3.8pt}
\renewcommand{\arraystretch}{1.12}
\resizebox{\columnwidth}{!}{%
\begin{tabular}{lcccc}
\toprule
Variant & NE$\downarrow$ & SR$\uparrow$ & OSR$\uparrow$ & SPL$\uparrow$ \\
\midrule
\textbf{Full} & \textbf{38.97} & \textbf{37.77} & \textbf{59.63} & \textbf{27.85} \\
w/o Hypothesis & 54.21 & 29.49 & 48.31 & 19.51 \\
w/o Exploration Memory & 47.35 & 33.65 & 53.49 & 22.57 \\
w/o Consistency Verification & 46.13 & 32.91 & 55.70 & 24.36 \\
w/o Active Evidence Seeking & 52.68 & 30.62 & 49.33 & 20.37 \\
w/o Adversarial Debate & 42.18 & 35.79 & 59.33 & 26.52 \\
\bottomrule
\end{tabular}}
\vspace{-2mm}
\end{table}

\noindent\textbf{Ablation Study.}
Table~\ref{tab:citynav_ablation} evaluates the contributions of key components in GeoReasoner. Removing geospatial hypotheses (w/o Hypothesis) causes the largest degradation, increasing NE by 15.24\,m and reducing SR and SPL by 8.28 and 8.34 points, respectively. This demonstrates the importance of constraining the search space before exploration.

Without exploration memory (w/o Exploration Memory), SR and SPL decrease by 4.12 and 5.28 points, highlighting the importance of accumulating evidence for informed exploration. Disabling active evidence seeking (w/o Active Evidence Seeking) leads to substantial drops of 7.15 and 10.30 points in SR and OSR, respectively, demonstrating the effectiveness of targeted observations in resolving spatial ambiguity and locating the goal.

For goal verification, replacing adversarial debate with direct MLLM assessment (w/o Adversarial Debate) reduces SR by 1.98 points, while removing consistency verification (w/o Consistency Verification) results in a larger decline of 4.86 points. The latter also decreases OSR by 3.93 points, highlighting the role of consistency verification in resolving ambiguous hypotheses and improving goal localization.

\begin{figure}[ht]
    \centering
    \includegraphics[width=\linewidth]{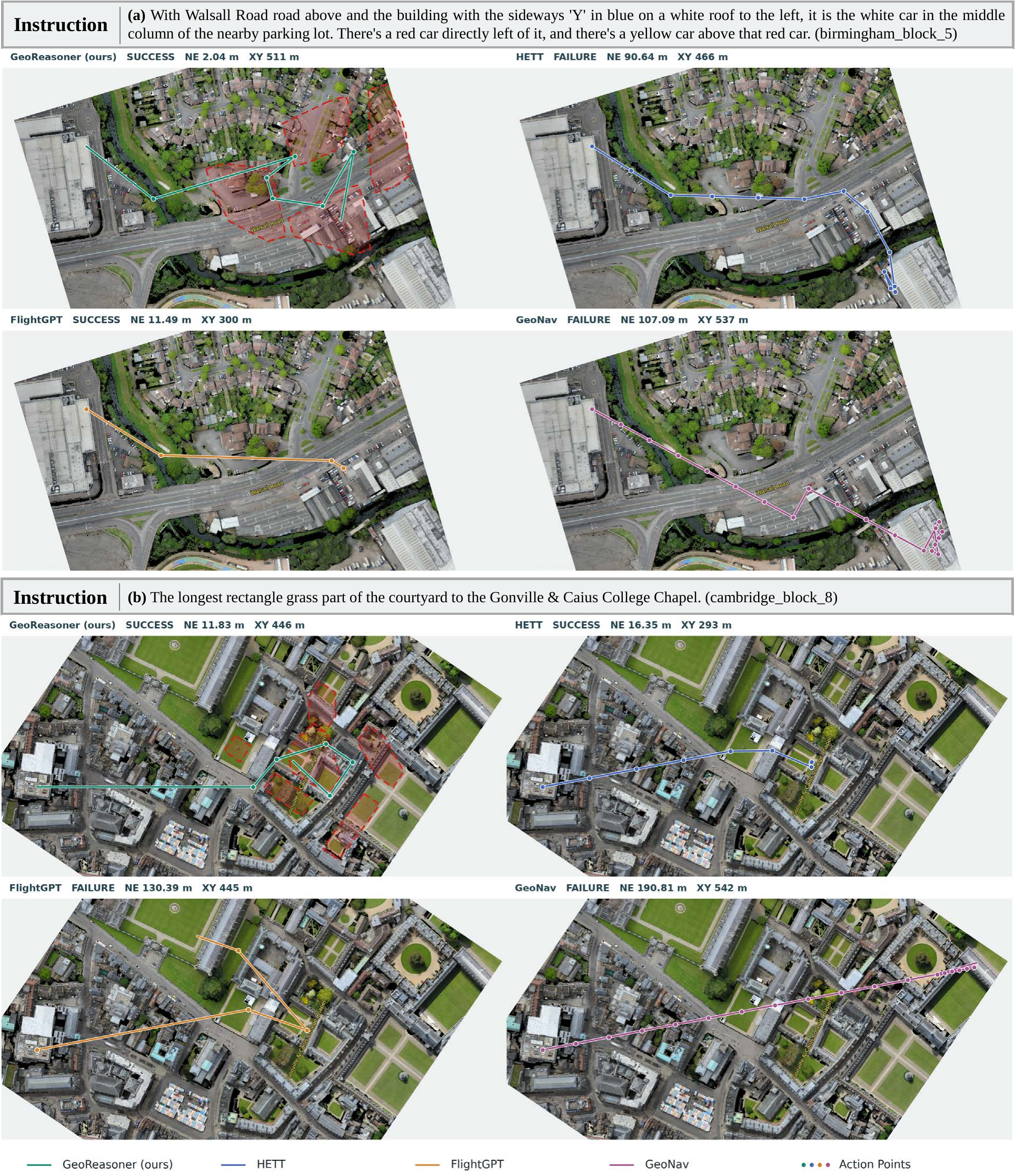}
    \caption{Qualitative comparison on CityNav. Recorded trajectories of GeoReasoner and three other methods in (a) fine-grained vehicle localization and (b) landmark-guided goal localization.}
    \label{vis}
\end{figure}
\noindent\textbf{Qualitative Results.}
Fig. \ref{vis} compares navigation trajectories in two challenging scenarios involving ambiguous visual targets and complex spatial relationships. In (a), GeoReasoner leverages road and building layouts to narrow the search region, then actively distinguishes the target white car from similar vehicles using spatial relations to nearby cars (NE: 2.04 m). In (b), identifying the correct rectangular grass region among similar courtyards is challenging, while GeoReasoner combines geospatial priors with targeted exploration and consistency verification to achieve an NE of 11.83 m. Unlike baselines that may follow misleading spatial cues, GeoReasoner effectively resolves spatial ambiguity through active evidence acquisition, enabling more accurate and reliable localization.

\section{Conclusion}
\label{sec:conclusion}
We present GeoReasoner, which advances aerial VLN beyond reactive search through geospatial-guided hypothesis generation, active evidence seeking, and consistency-aware verification. Experiments demonstrate superior zero-shot performance and robustness across navigation difficulties.

\section*{LLM Usage Statement}
We used Qwen3.8-Flash for multimodal reasoning in our proposed framework and ChatGPT (OpenAI) for language polishing and manuscript refinement. All research design, experimental results, and
scientific claims were verified by the authors.

{
    \small
    \bibliographystyle{ieeenat_fullname}
    \bibliography{main}

@String(CVPR= {IEEE Conf. Comput. Vis. Pattern Recog.})

@String(ICCV= {Int. Conf. Comput. Vis.})

@String(AAAI = {AAAI})

@String(CVPR  = {CVPR})

@String(ICCV  = {ICCV})

@inproceedings{liu2023aerialvln,
  title={Aerialvln: Vision-and-language navigation for uavs},
  author={Liu, Shubo and Zhang, Hongsheng and Qi, Yuankai and Wang, Peng and Zhang, Yanning and Wu, Qi},
  booktitle={Proceedings of the IEEE/CVF International Conference on Computer Vision},
  pages={15384--15394},
  year={2023}
}

@inproceedings{lee2025citynav,
  title={Citynav: A large-scale dataset for real-world aerial navigation},
  author={Lee, Jungdae and Miyanishi, Taiki and Kurita, Shuhei and Sakamoto, Koya and Azuma, Daichi and Matsuo, Yutaka and Inoue, Nakamasa},
  booktitle={Proceedings of the IEEE/CVF International Conference on Computer Vision},
  pages={5912--5922},
  year={2025}
}

@inproceedings{anderson2018vision,
  title={Vision-and-language navigation: Interpreting visually-grounded navigation instructions in real environments},
  author={Anderson, Peter and Wu, Qi and Teney, Damien and Bruce, Jake and Johnson, Mark and S{\"u}nderhauf, Niko and Reid, Ian and Gould, Stephen and Van Den Hengel, Anton},
  booktitle={Proceedings of the IEEE conference on computer vision and pattern recognition},
  pages={3674--3683},
  year={2018}
}

@inproceedings{ku2020room,
  title={Room-across-room: Multilingual vision-and-language navigation with dense spatiotemporal grounding},
  author={Ku, Alexander and Anderson, Peter and Patel, Roma and Ie, Eugene and Baldridge, Jason},
  booktitle={Proceedings of the 2020 Conference on Empirical Methods in Natural Language Processing (EMNLP)},
  pages={4392--4412},
  year={2020}
}

@inproceedings{wang2025openbench,
  title={OpenBench: A new benchmark and baseline for semantic navigation in smart logistics},
  author={Wang, Junhui and Huo, Dongjie and Xu, Zehui and Shi, Yongliang and Yan, Yimin and Wang, Yuanxin and Gao, Chao and Qiao, Yan and Zhou, Guyue},
  booktitle={2025 IEEE International Conference on Robotics and Automation (ICRA)},
  pages={16202--16208},
  year={2025},
  organization={IEEE}
}

@inproceedings{doschl2024say,
  title={Say-REAPEx: An LLM-Modulo UAV online planning framework for search and rescue},
  author={D{\"o}schl, Bj{\"o}rn and Kiam, Jane Jean},
  booktitle={2nd CoRL Workshop on Learning Effective Abstractions for Planning},
  year={2024}
}

@article{huang2021decentralized,
  title={Decentralized autonomous navigation of a UAV network for road traffic monitoring},
  author={Huang, Hailong and Savkin, Andrey V and Huang, Chao},
  journal={IEEE Transactions on Aerospace and Electronic Systems},
  volume={57},
  number={4},
  pages={2558--2564},
  year={2021},
  publisher={IEEE}
}

@inproceedings{zhang2025citynavagent,
  title={Citynavagent: Aerial vision-and-language navigation with hierarchical semantic planning and global memory},
  author={Zhang, Weichen and Gao, Chen and Yu, Shiquan and Peng, Ruiying and Zhao, Baining and Zhang, Qian and Cui, Jinqiang and Chen, Xinlei and Li, Yong},
  booktitle={Proceedings of the 63rd Annual Meeting of the Association for Computational Linguistics (Volume 1: Long Papers)},
  pages={31292--31309},
  year={2025}
}

@article{xu2026geonav,
  title={GeoNav: Empowering MLLMs with dual-scale geospatial reasoning for language-goal aerial navigation},
  author={Xu, Haotian and Hu, Yue and Gao, Chen and Zhu, Zhengqiu and Zhao, Yong and Yin, Quanjun},
  journal={Pattern Recognition},
  pages={113365},
  year={2026},
  publisher={Elsevier}
}

@inproceedings{ning2026lookasidevln,
  title={LookasideVLN: direction-aware aerial vision-and-language navigation},
  author={Ning, Yuwei and Zhao, Ganlong and Qin, Yipeng and Liu, Si and Liu, Yang and Lin, Liang and Li, Guanbin},
  booktitle={Proceedings of the IEEE/CVF Conference on Computer Vision and Pattern Recognition},
  pages={32441--32450},
  year={2026}
}

@article{chen2021hamt,
  title={History aware multimodal transformer for vision-and-language navigation},
  author={Chen, Shizhe and Guhur, Pierre-Louis and Schmid, Cordelia and Laptev, Ivan},
  journal={Advances in neural information processing systems},
  volume={34},
  pages={5834--5847},
  year={2021}
}

@inproceedings{chen2022duet,
  title={Think global, act local: Dual-scale graph transformer for vision-and-language navigation},
  author={Chen, Shizhe and Guhur, Pierre-Louis and Tapaswi, Makarand and Schmid, Cordelia and Laptev, Ivan},
  booktitle={2022 IEEE/CVF Conference on Computer Vision and Pattern Recognition (CVPR)},
  pages={16516--16526},
  year={2022},
  organization={IEEE}
}

@inproceedings{zhou2024navgpt2,
  title={Navgpt-2: Unleashing navigational reasoning capability for large vision-language models},
  author={Zhou, Gengze and Hong, Yicong and Wang, Zun and Wang, Xin Eric and Wu, Qi},
  booktitle={European Conference on Computer Vision},
  pages={260--278},
  year={2024},
  organization={Springer}
}

@article{guo2026awarevln,
  title={AwareVLN: Reasoning with Self-awareness for Vision-Language Navigation},
  author={Guo, Wenxuan and Xu, Xiuwei and Liu, Yichen and Li, Xiangyu and Yin, Hang and Chen, Huangxing and Zheng, Wenzhao and Feng, Jianjiang and Zhou, Jie and Lu, Jiwen},
  journal={arXiv preprint arXiv:2605.22816},
  year={2026}
}

@article{xue2026profocus,
  title={ProFocus: Proactive Perception and Focused Reasoning in Vision-and-Language Navigation},
  author={Xue, Wei and Li, Mingcheng and Wu, Xuecheng and Tang, Jingqun and Yang, Dingkang and Zhang, Lihua},
  journal={arXiv preprint arXiv:2603.05530},
  year={2026}
}

@inproceedings{cai2025flightgpt,
  title={Flightgpt: Towards generalizable and interpretable uav vision-and-language navigation with vision-language models},
  author={Cai, Hengxing and Dong, Jinhan and Tan, Jingjun and Deng, Jingcheng and Li, Sihang and Gao, Zhifeng and Wang, Haidong and Su, Zicheng and Sumalee, Agachai and Zhong, Renxin},
  booktitle={Proceedings of the 2025 Conference on Empirical Methods in Natural Language Processing},
  pages={6670--6687},
  year={2025}
}

@inproceedings{fan2026htnav,
  title={HTNav: A Hybrid Navigation Framework with Tiered Structure for Urban Aerial Vision-and-Language Navigation},
  author={Chengjie Fan and
                  Cong Pan and
                  Zijian Liu and
                  Ningzhong Liu and
                  Jie Qin},
  booktitle    = {{IEEE/CVF} Conference on Computer Vision and Pattern Recognition},
  pages        = {10976--10985},
  year={2026}
}

@inproceedings{ding2026history,
  title={History-enhanced two-stage transformer for aerial vision-and-language navigation},
  author={Ding, Xichen and Gao, Jianzhe and Pan, Cong and Wang, Wenguan and Qin, Jie},
  booktitle={Proceedings of the AAAI Conference on Artificial Intelligence},
  volume={40},
  number={22},
  pages={18225--18233},
  year={2026}
}

@inproceedings{qi2024shapellm,
  title={Shapellm: Universal 3d object understanding for embodied interaction},
  author={Qi, Zekun and Dong, Runpei and Zhang, Shaochen and Geng, Haoran and Han, Chunrui and Ge, Zheng and Yi, Li and Ma, Kaisheng},
  booktitle={European Conference on Computer Vision},
  pages={214--238},
  year={2024},
  organization={Springer}
}

@inproceedings{ma2025spatialllm,
  title={Spatialllm: A compound 3d-informed design towards spatially-intelligent large multimodal models},
  author={Ma, Wufei and Ye, Luoxin and de Melo, Celso M and Yuille, Alan and Chen, Jieneng},
  booktitle={2025 IEEE/CVF Conference on Computer Vision and Pattern Recognition (CVPR)},
  pages={17249--17260},
  year={2025},
  organization={IEEE}
}

@inproceedings{yang2025thinking,
  title={Thinking in space: How multimodal large language models see, remember, and recall spaces},
  author={Yang, Jihan and Yang, Shusheng and Gupta, Anjali W and Han, Rilyn and Fei-Fei, Li and Xie, Saining},
  booktitle={2025 IEEE/CVF Conference on Computer Vision and Pattern Recognition (CVPR)},
  pages={10632--10643},
  year={2025},
  organization={IEEE}
}

@inproceedings{liu2025coarse,
  title={Coarse correspondences boost spatial-temporal reasoning in multimodal language model},
  author={Liu, Benlin and Dong, Yuhao and Wang, Yiqin and Ma, Zixian and Tang, Yansong and Tang, Luming and Rao, Yongming and Ma, Wei-Chiu and Krishna, Ranjay},
  booktitle={2025 IEEE/CVF Conference on Computer Vision and Pattern Recognition (CVPR)},
  pages={3783--3792},
  year={2025},
  organization={IEEE}
}

@inproceedings{wang2025spatial457,
  title={Spatial457: A diagnostic benchmark for 6d spatial reasoning of large multimodal models},
  author={Wang, Xingrui and Ma, Wufei and Zhang, Tiezheng and de Melo, Celso M and Chen, Jieneng and Yuille, Alan},
  booktitle={2025 IEEE/CVF Conference on Computer Vision and Pattern Recognition (CVPR)},
  pages={24669--24679},
  year={2025},
  organization={IEEE}
}

@inproceedings{daxberger2025mmspatial,
  title={Mm-spatial: Exploring 3d spatial understanding in multimodal llms},
  author={Daxberger, Erik and Wenzel, Nina and Griffiths, David and Gang, Haiming and Lazarow, Justin and Kohavi, Gefen and Kang, Kai and Eichner, Marcin and Yang, Yinfei and Dehghan, Afshin and others},
  booktitle={2025 IEEE/CVF International Conference on Computer Vision (ICCV)},
  pages={7395--7408},
  year={2025},
  organization={IEEE}
}

@article{hu2026ringmo,
  title={RingMo-Agent: A unified remote sensing foundation model for multi-platform and multi-modal reasoning},
  author={Hu, Huiyang and Wang, Peijin and Feng, Yingchao and Wei, Kaiwen and Yin, Wenxin and Diao, Wenhui and Wang, Mengyu and Bi, Hanbo and Kang, Kaiyue and Ling, Tong and others},
  journal={IEEE Transactions on Pattern Analysis and Machine Intelligence},
  year={2026},
  publisher={IEEE}
}

@article{qiu2026design,
  title={On the design of qwen3. 8-next architecture: Evaluation, efficiency, and training stability},
  author={Qiu, Zihan and Wang, Zekun and Li, Xiao and Li, Yanpeng and Xu, Yang and Wang, Yixuan and Zhang, Huaqing and Men, Rui and Mao, Bochao and Zhang, Chengruidong and others},
  journal={arXiv preprint arXiv:2608.30320},
  year={2026}
}

@inproceedings{carion2026sam,
  title={Sam 3: Segment anything with concepts},
  author={Carion, Nicolas and Gustafson, Laura and Hu, Yuan-Ting and Debnath, Shoubhik and Hu, Ronghang and Suris Coll-Vinent, Didac and Ryali, Chaitanya and Alwala, Kalyan Vasudev and Khedr, Haitham and Huang, Andrew and others},
  booktitle={International conference on learning representations},
  volume={2026},
  pages={138846--138923},
  year={2026}
}

@incollection{applegate2011traveling,
  title={The traveling salesman problem: a computational study},
  author={Applegate, David L and Bixby, Robert E and Chv{\'a}tal, Va{\v{s}}ek and Cook, William J},
  booktitle={The traveling salesman problem},
  year={2011},
  publisher={Princeton university press}
}

@article{liu2026dronenav,
  title={DroneNav: Unified text-visual representation and structured spatial reasoning for robust UAV vision-and-language navigation},
  author={Liu, Fangming and Li, Guohua and Zou, Linfeng and Chen, Yadong and Cheng, Peng},
  journal={Neurocomputing},
  pages={133492},
  year={2026},
  publisher={Elsevier}
}

@article{liu2026gpod,
  title={GPOD: Geographic Priors and Object Detection for Candidate-Guided Target Localization in City-Scale UAV Vision-and-Language Navigation},
  author={Liu, Yuze and Xu, Changming and Xiao, Kewen and Wu, Yuhua and Li, Ziyu},
  journal={Drones},
  volume={10},
  number={6},
  pages={458},
  year={2026},
  publisher={MDPI}
}

@inproceedings{krantz2020beyond,
  title={Beyond the nav-graph: Vision-and-language navigation in continuous environments},
  author={Krantz, Jacob and Wijmans, Erik and Majumdar, Arjun and Batra, Dhruv and Lee, Stefan},
  booktitle={European Conference on Computer Vision},
  pages={104--120},
  year={2020},
  organization={Springer}
}
}

% WARNING: do not forget to delete the supplementary pages from your submission 
% \input{sec/X_suppl}

\end{document}